\documentclass[10pt,twocolumn,letterpaper]{article}

\usepackage{cvpr}
\usepackage{times}
\usepackage{epsfig}
\usepackage{graphicx}
\usepackage{amsmath}
\usepackage{amssymb}
\usepackage{authblk}
\usepackage{color}
\usepackage{tabularx}
\usepackage{caption}

\usepackage[pagebackref=true,breaklinks=true,letterpaper=true,colorlinks,bookmarks=false]{hyperref}

\cvprfinalcopy 

\ifcvprfinal\fi
\makeatletter
\newcommand{\leqnomode}{\tagsleft@true}
\newcommand{\reqnomode}{\tagsleft@false}
\makeatother

\newcommand*\samethanks[1][\value{footnote}]{\footnotemark[#1]}

\makeatletter
\renewcommand\AB@affilsepx{, \protect\Affilfont}
\makeatother
\title{Depth Assisted Full Resolution Network for Single Image-based View Synthesis}
\author[1,a]{Xiaodong Cun\thanks{These two authors contributed equally}}
\author[2,b]{Feng Xu\samethanks}
\author[1,c]{Chi-Man Pun\thanks{Corresponding Author}}
\author[1,3,d]{Hao Gao}

\affil[1]{University of Macau}
\affil[2]{Tsinghua University}
\affil[3]{Nanjing University of Posts and Telecommunications}

\affil[a]{\tt\small\href{mailto:mb55411@umac.mo}{mb55411@umac.mo}}
\affil[b]{\tt\small\href{mailto:feng-xu@tsinghua.edu.cn}{feng-xu@tsinghua.edu.cn}}
\affil[c]{\tt\small\href{mailto:cmpun@umac.mo}{cmpun@umac.mo}}
\affil[d]{\tt\small\href{mailto:tsgaohao@gmail.com}{tsgaohao@gmail.com}}

\begin{document}

\maketitle

\begin{abstract}
Researches in novel viewpoint synthesis majorly focus on interpolation from multi-view input images. In this paper, we focus on a more challenging and ill-posed problem that is to synthesize novel viewpoints from one single input image. To achieve this goal, we propose a novel deep learning-based technique. We design a full resolution network that extracts local image features with the same resolution of the input, which contributes to derive high resolution and prevent blurry artifacts in the final synthesized images. We also involve a pre-trained depth estimation network into our system, and thus 3D information is able to be utilized to infer the flow field between the input and the target image. Since the depth network is trained by depth order information between arbitrary pairs of points in the scene, global image features are also involved into our system. Finally, a synthesis layer is used to not only warp the observed pixels to the desired positions but also hallucinate the missing pixels with recorded pixels. Experiments show that our technique performs well on images of various scenes, and outperforms the state-of-the-art techniques.
\end{abstract}

\vspace{-2mm}
\section{Introduction}
Synthesizing images of novel viewpoints is widely investigated in computer vision and graphics.
Most works in this topic focus on using multi-view images to synthesize viewpoints in-between~\cite{Leimkuehler2017TVCG,Efrat2016,DidykSFDM2013,Kellnhofer2017,Binolf2015,Ji_2017_CVPR,LearningViewSynthesis}.
In this paper, we consider extrapolation, and we take a step further to do extrapolation from one single input image.
This technique is quite useful for many applications, such as multi-view rendering in virtual reality, light field reconstruction from less recording\cite{LearningViewSynthesis}, and image post-processing like image refocusing.
However, this task is very challenging for two major reasons.
First, some parts of the scene may be not observed in the input viewpoint but are required for novel ones.
Second, 3D information is lacking for single view input but is crucial to determine pixel changes between viewpoints.
Although very challenging, we observe that human brains are always able to imagine novel viewpoints.
The reason is that human brains have learned in our daily lives to understand the depth order of objects in a scene~\cite{NIPS2016_6489} and infer what the scene looks like when viewing from another viewpoint.



\begin{figure*}
 	\centering
    \includegraphics[width=\textwidth]{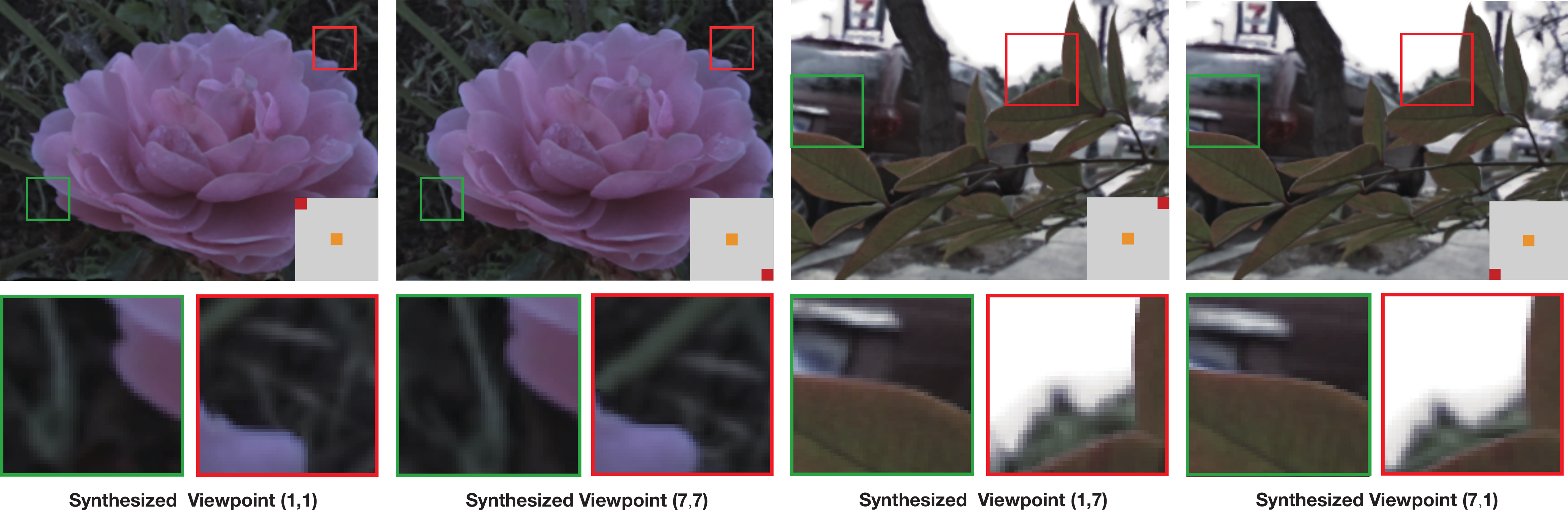}
    \captionof{figure}{\color{black}We propose a deep learning based method to synthesize surrounding novel views from one input center view. We mark the coordinates of the center view as a yellow dot and a synthesized viewpoint as a red dot in a gray square, indicating the relative position of the viewpoints. Here we show our results on Flower dataset (left two) and natural image dataset (right two) at four extreme viewpoint positions. The zoomed-in regions contain both foreground and background whose relative positions are changed according to the changes in viewpoint.}
    \label{fig:0}
\end{figure*}

Inspired by human brains, we use deep neural networks to learn to synthesize novel viewpoint from a light field dataset (A result is shown in Figure.~\ref{fig:0}).
We believe that for novel viewpoint synthesis, both global and local image features are important. And our key observation is that after modeling the process as two steps: depth prediction and depth-based image warping, the extraction of the two kinds of features can be decoupled.
Depth estimation from a single image is ill-posed, so global high-level image features are required to tackle the problem.
But given the depth, only local image warping is required to synthesize the final result, and local depth and color information are enough to determine the warping.
Based on this observation, we focus on the two kinds of features in the two steps respectively.

First, we explicitly estimate depth information from images by global high-level image features.
As learning global features requires a large dataset to cover sufficient variations and the current light field dataset is small and lack of coverage, we leverage an existing large image dataset (421k images) with labeled depth orders to pre-train a depth prediction network~\cite{NIPS2016_6489}.
Then, with the good depth information, our view synthesis network is further trained to extract local features directly from a light field dataset. As local features do not have as much variations as global ones, current light field datasets are relatively sufficient, which is also demonstrated in our experiments.

However, the network design is not trivial to extract important local features for view synthesis.
Existing deep CNNs \cite{he2015,Simonyan14c,he2017mask,zhou2016view,qifeng2017ICCV,long2015fully,yu2015multi} majorly focus on extracting global high level features, which have two key drawbacks.
First, global features are usually invariant to spatial transformations (scale, translation, and rotation). It is not desired for novel view synthesis which needs to delicately change the orientations of objects on the 2D image domain.
Second, global features are expected to be invariant to local details. It is not desired either because local details need to be correctly modified to guarantee a reasonable synthesis of novel viewpoints.
Based on these observations, we propose a full resolution network (whose all layers are with the same size of the input image) to encode content orientations and images details, which benefit to achieve high quality and high-resolution view synthesis.

Recently, a concurrent work~\cite{pratul2017lightField} also uses deep learning to synthesize novel viewpoints from a single image for flowers, where light field images of flowers are used to train a network which first infers per-pixel depth values and then synthesize a 4D light field by dilation convolutions.
Compared with this work, we have two major differences.
First, our network for depth prediction is trained on plenty of images from various scenes, instead of a small set of light field images, so the depth prediction is expected to have better generalization capability.
Second, since the depth is obtained by a pre-trained network, the following synthesizing network only needs to combine local image features with the depth to infer the local image warping.
Thus our full resolution network focuses on local features and does not need dilation convolutions to enlarge receptive fields, where gridding artifacts are inevitable. 

We summarize our main contributions as follows:
\begin{itemize}
\item We propose a depth assisted full resolution network to synthesize user-desired viewpoint from a single natural image, which achieves the state-of-the-art performance over the existing techniques.
\item We leverage a large image dataset for depth prediction, which breaks the limitation caused by the small size of the current light field datasets.
\item We propose a full resolution network that extracts local image features to warp local image details in the synthesis.

\end{itemize}

\section{Related work}


\textbf{View synthesis for scenes} is a popular problem in both computer vision and computer graphics. Usually, synthesizing novel views requires estimating disparity from multiple input images, and then the synthesis is performed by warping input images with the disparity. \cite{DidykSFDM2013,Efrat2016,Kellnhofer2017} synthesize automultiscopic images from stereo images. \cite{Binolf2015} presents a disparity based method to reconstruct light field from micro-image pairs. However, in our problem, the surrounding views need to be predicted by a single image.

Recently, deep learning based methods have been utilized in novel view synthesis. {\color{black}\cite{flynn2016deepstereo} is inspired by traditional plane sweep algorithms and uses multiple input images to learn the target image of a novel viewpoint.} \cite{jaderberg2015spatial} learns a set of transformation parameters describing the relationship between the input image and the target image. \cite{xie2016deep3d} tries to synthesize stereo image pairs from a single input image. It just generates the corresponding right image from the left image, without the controllability of the coordinates of the target image. \cite{Ji_2017_CVPR} presents a convolutional network based method to morph a novel view from a stereo image pair. \cite{LearningViewSynthesis,EPICNN17} synthesize novel views from four corner images, which focus on reconstructing the light field from multiple input images. \cite{pratul2017lightField} trains a network from flower dataset to reconstruct the light field images from one single image.

\textbf{View synthesis for objects} is another related topic in computer vision. The 3D structure of a single object can be predicted by a single input image. \cite{dosovitskiy2015learning} tries to synthesize an object at novel views from a single input image by neural networks directly. \cite{zhou2016view} extends this method by estimating an appearance flow. By considering the relationship between different viewpoints of the same object, \cite{yang2015weakly} proposes a network deriving from recurrent neural network(RNN) for chair synthesis. Most recently, a generative adversarial network \cite{goodfellow2014generative} is proposed by \cite{tvsncvpr2017} to predict novel views from a single input image. Although these methods can generate novel viewpoints that are quite different from the input, they just work well on simple objects, such as cars and chairs.

\textbf{Single view depth estimation} reconstructs per-pixel depth of an input image, which is an important information in the view synthesis task. \cite{saxena2008make3d} tries to predict depth by learning algorithms. Deeper networks have also been tested by \cite{eigen2015predicting,laina2016deeper}. These methods have limited accuracy in natural scenes because scene variations are restricted by the coverage of their training datasets. Inspired by depth based image rendering, unsupervised methods \cite{monodepth17,garg2016unsupervised} based on stereo constraints have shown great potential in depth estimation. However, the camera parameters are required for these techniques. Varying from previous methods, \cite{NIPS2016_6489} proposes a relative depth-based framework to estimate depth for unconstrained images (``in the wild"). It leverages a large dataset containing 421k images and performs well in deciding the relative position of objects in the scenes.

\vspace{-3mm}

\section{Methods}
 Given a single image $I_p$ at the center viewpoint $p$ and the position of a novel viewpoint $q$, our goal is to synthesize a novel image $I_q$ at the viewpoint $q$. This problem can be formulated as:
\reqnomode
\begin{align}\label{eq:full1}
 I_q=\Phi(I_p,q),
\end{align}
where $\Phi$ is a function which defines the relationship between $I_q$ and $I_p$. The view point position $p$ and $q$ are represented in a 2D coordinate system which centers at the viewpoint of $p$ and is orthogonal to the viewing direction. 

We propose a depth assisted full resolution network for novel view synthesis. In our pipeline, we train the network to infer a flow field describing the relationship between the input image and the novel view. Then we warp the input image by the flow field. Following this strategy, we can reformulate Equation. \ref{eq:full1} as follows:
\reqnomode
\begin{align}\label{eq:full2}
F_q=\Phi_d(I_p,q), \\
I_q=\Phi_{w}(I_p,F_q),
\end{align}
where $\Phi_d$ describes our network which predicts a pixel-level flow field $F_q$ for the novel view $q$. $\Phi_w$ is the warping method that warps the input image $I_p$ with the flow field $F_q$ on viewpoint $q$.

\vspace{-1.5mm}
\subsection{Depth Assisted Full Resolution Network}

\begin{figure}
\setlength{\abovecaptionskip}{0.cm}
\setlength{\belowcaptionskip}{-0.4cm}
\centering
\includegraphics[width=0.9\columnwidth]{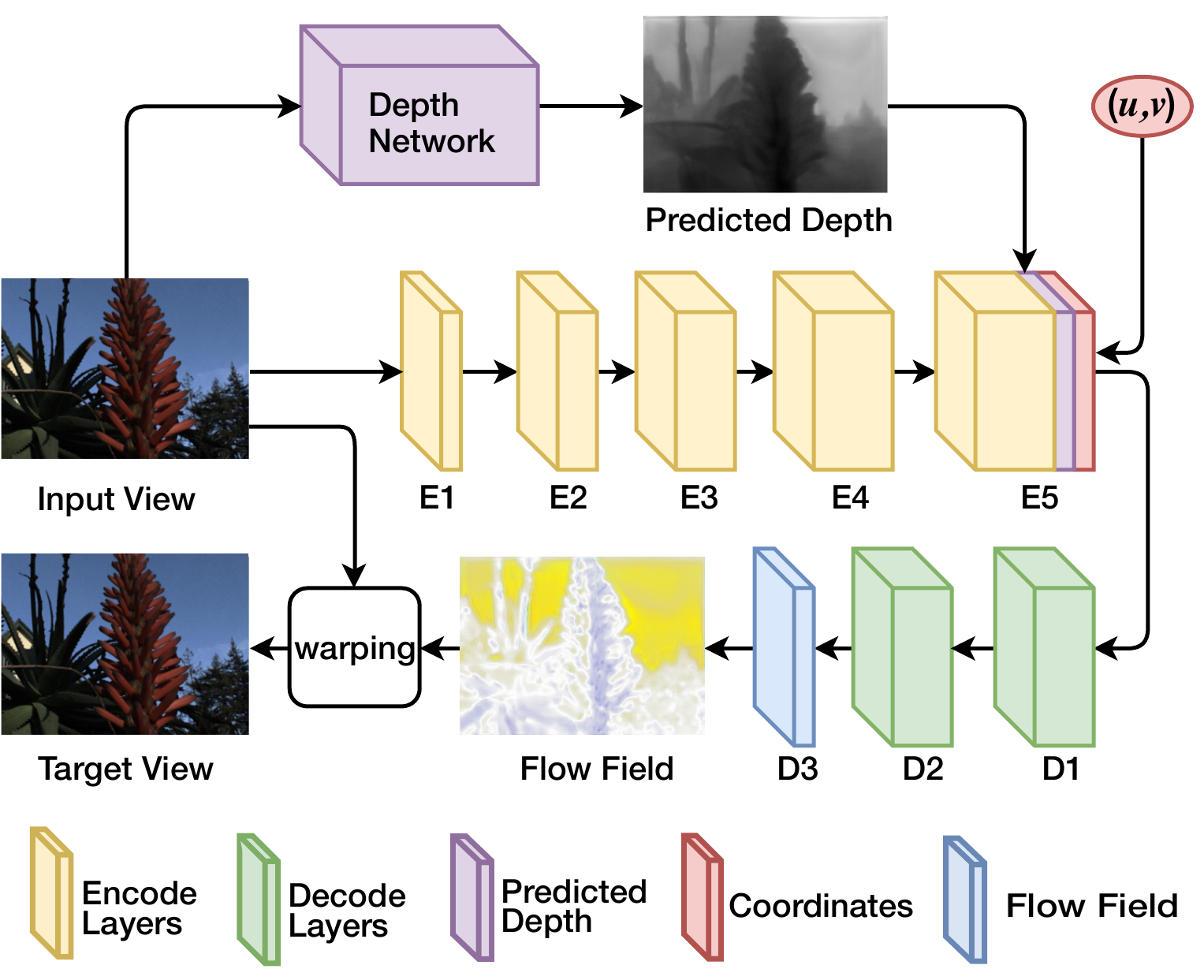}
\caption{Network Structure. A full resolution network is proposed for our problem. A depth, estimated by a pre-trained depth predictor, is added as a feature map in the network, as well as $(u,v)$ denoting the coordinates of the target viewpoint. Then, a flow field is decoded from the combined feature and a warping layer is added at the end of the network to synthesize the final target view.}\label{fig:1}
\end{figure}

In this section, we illustrate our depth assisted full resolution network for novel view synthesis (Figure.~\ref{fig:1}). The encoder part of our full resolution network extracts important local features from the input image. Then our depth predictor, which is pre-trained on a large image dataset by exploring global image information, estimates a depth map of the input image. Next the local features and the depth are fed into our decoder, as well as a 2-channel map indicating the position of the target viewpoint. Finally, our decoder translates the combined features into a warping field to synthesize the final target image.


\vspace{-1mm}
\subsubsection{Encoder}\label{sec:encoder}

The encoder is designed for extracting local features of the input image. Following the method in \cite{LearningViewSynthesis}, the encoder network is a series of convolution kernels with different convolution kernel sizes but generating features of the same resolution with the input image. A Rectified Linear Units (ReLU) layer is added after each convolution layers. These features will be used to rebuild the final transformed image. {\color{black}Notice that we do not use pooling or batch normalization for reducing the parameters and resolutions of the network as they may be harmful to encode content orientations and image details as discussed before.}
\vspace{-1mm}
\subsubsection{Feature Connection} \label{sec::connection}

As shown in Figure. \ref{fig:1}, we add the predicted relative depth, estimated by \cite{NIPS2016_6489}, as a feature of the input image. 
\cite{NIPS2016_6489} trains a depth prediction network from the labeled depth ranking of pixel pairs on one image. So the output indicates the relative depth of the input images. 421k images, which are gathered from Flickr and marked by crowdsourcing with the relative depth ordering of two random pixels, are used for training the network. We only take the forward output of this network for extracting the depth of our input image because we do not have the ground truth depth for backward training.

There are four main advantages for using this depth feature: Firstly, depth is a closely relevant feature of our flow field. The relationship of depth $Z$ and disparity $D$ between an input image and a novel view can be written as \cite{xie2016deep3d}:
\begin{align}\label{eq:depth}
D = \frac {B(Z-f)} {Z},
\end{align}
where $B$ is the absolute distance between two viewpoints and $f$ is the focus. There is also a clear relationship between the disparity $D_q$ of novel view $q$ and our flow field $F_q$:
\begin{align}\label{eq:depth2}
 F_q(s) =  ( D_q(s) \times \Delta u , D_q(s) \times \Delta v ),
\end{align}
where $\Delta u$, $\Delta v$ are the differences of the viewpoint coordinates in $u$, $v$ direction, respectively. According to Equation. \ref{eq:depth} and Equation. \ref{eq:depth2}, depth information is very important for estimating flow field.
Secondly, \cite{NIPS2016_6489} predicts the relative depth of an image, which gives a more clear relative position relationship between objects than other methods \cite{eigen2015predicting,laina2016deeper}. And it is important to generate head motion parallax for viewpoint transitions. Thirdly, the network for predicting the depth has been trained by information (depth orders) of two largely apart pixels, so large perception field is implicitly considered by our network by involving the depth. As the full resolution network preserves local features, we have collected both local and global information for the final synthesis. Finally, the dataset for training the depth predictor is very large in size and covers a large amount of nature scenes, so it is important for the generalization capability of our system.

Besides the depth image connected to the network as a feature layer in the end of the encoder part, the 2D coordinates $(u, v)$ of the novel view is also added as two layer features which have the same size as the input image~\cite{LearningViewSynthesis}. This is for giving the viewpoint information of the target to the network.


\vspace{-1mm}
\subsubsection{Decoder} \label{sec:decoder}

The network in this part estimates the dense flow for all pixels. It has been demonstrated in several previous works \cite{zhou2016view,tvsncvpr2017} that synthesizing the flow from the input view to novel views is better than generating the novel views by networks directly. Notice that {\color{black} as we use backward interpolate method}, the flow field is also used to handle the occlusion regions which is not visible in the input. In this case, the flow does not stand for real correspondences, but is only used to hallucinate the occlusion regions, and is also learned in our method. The network in this decoder part contains four convolution layers. The first three are followed by ReLU layers, while the last one is followed by a $Tanh$ layer.
\vspace{-1mm}
\subsubsection{Flow based warping}\label{sec:warp}

Following the idea of appearance flow \cite{zhou2016view} and the spatial transform network \cite{ranjan2016optical}, we apply flow based warping method for synthesizing the final image. There is a clear mathematical relationship between the predicted flow fields and novel view images. For every pixel $s$ in one novel view image, its pixel value can be expressed as:
\begin{align}
  I_q(s) = I_p[ s + F_q(s) ],
	\label{eq:warp}
\end{align}
where $F_q(s)$ is the two-dimensional flow which is the output of our neural network. Here, a backward warping is utilized to transform the input image to the novel view as the flow is defined at pixel $s$ on the target view.
Since the warping function described in Equation. \ref{eq:warp} is differentiable, and the gradient can be calculated efficiently \cite{jaderberg2015spatial}, all the layers of our network are differentiable and the whole network can be trained end-to-end in a supervised manner.
The details of our network can be found in Table \ref{tab:1}.

\begin{table}[hbp!]
\setlength{\abovecaptionskip}{-0.0cm}
\setlength{\belowcaptionskip}{-0.3cm}
\centering
\caption{Network Structure}
\label{tab:1}
\begin{tabular}{lccc}
\hline
           &  convolution size  & input channel & output channel      \\ \hline
E1         & 	$7\times7$      & 3         	& 32          \\
E2         & 	$5\times5$      & 32         	& 64          \\
E3         & 	$3\times3$      & 64          	& 128          \\
E4         & 	$1\times1$      & 128         	& 192          \\
D1         & 	$3\times3$      & 195          	& 192          \\
D2         & 	$3\times3$		& 192 			& 128			\\	
D3         & 	$3\times3$		& 128 			& 64			\\	
D4         & 	$3\times3$ 		& 64 			& 2 			\\ \hline
\end{tabular}

\end{table}

\begin{figure*}[tb]
\setlength{\abovecaptionskip}{-0.0cm}
\setlength{\belowcaptionskip}{-0.3cm}
\centering
\includegraphics[width=\textwidth]{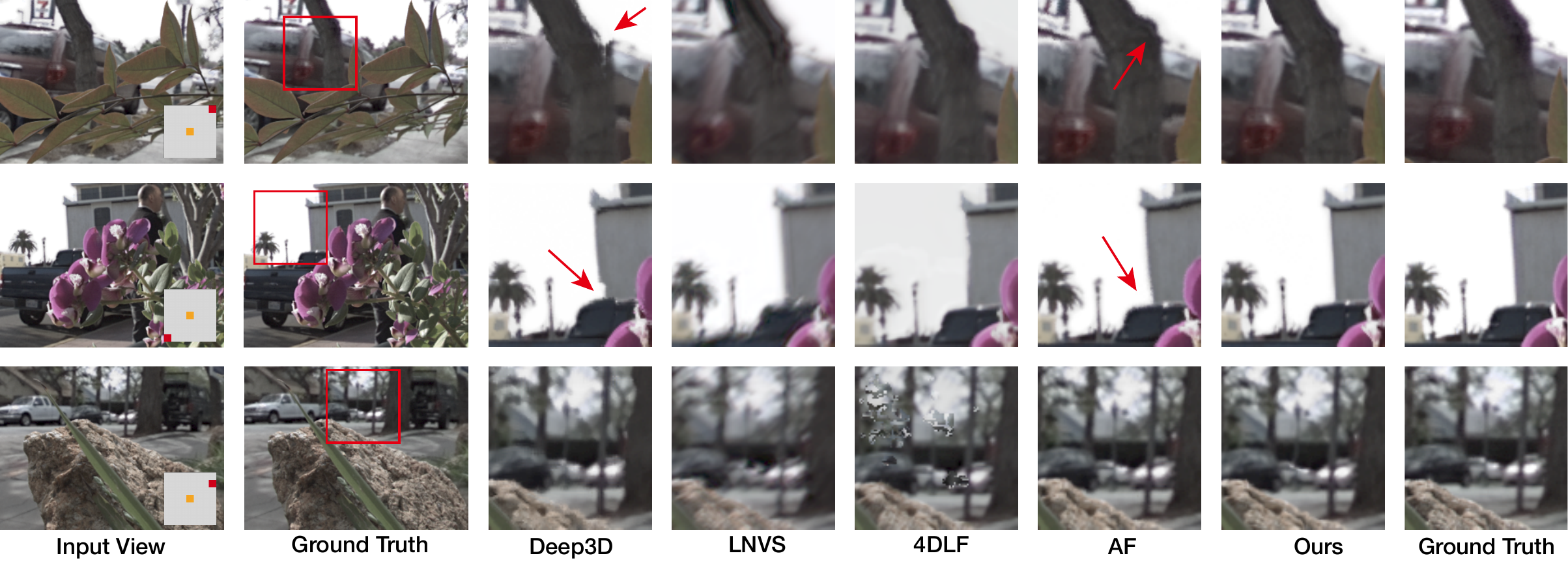}
\caption{Comparison with the state-of-art methods. We illustrate several selected patches on different viewpoints of different input images. Our method shows clear results and few artifacts than other alternative methods. All viewpoints results can be found in the supplementary video. }\label{fig:3}
\end{figure*}
\vspace{-1.5mm}
\subsection{Loss Function}
The objective function $C$ of our network can be written as:
\begin{align}
 C = L_1(I_q , \widehat{I}_q) + \alpha L_{TV}(F_{q})
\end{align}

The first part of the loss function is a traditional image reconstruction error($L_1$), which restricts the similarity between the result $I_q$ and the ground truth $\widehat{I_q}$. The second part of our loss function is a total variation regularization for the predicted flow field $F_{q}$. It is necessary to add the regularization to our method because the total variation constraint in the flow field $F_{q}$ will guarantee smoothness and produce high quantity results. we empirically set $\alpha = 0.001$ for all the experiments.

\vspace{-1.5mm}
\subsection{Training Details}

\subsubsection{Relative position for training}
Lytro Illum camera captures the light field of a scene by a regular microlens array.
Because of the distances among the viewpoints are much less than the distance between the camera and the scene object, we assume all the viewpoints are in a 2D $u-v$ plane.
In the training, we denote the position of the center viewpoint $p_{center}(u, v)$ as $[0, 0]$. And $p_{novel}(u, v)$ ranges in $[- 3, + 3]\times[- 3, + 3]$, accordingly.
To make full use of the dataset, all light field images will have a chance to be chosen as the center view, and the coordinates of other images are determined by their relative positions to the center image.
\subsubsection{Datasets and Parameters}
We have used two datasets for experiments and validates. One is a light field dataset from \cite{LearningViewSynthesis} (VS100 dataset). It contains 100 training images and 30 test images with the angular resolution of $8\times8$ . Diverse scenes, such as cars, flowers, and trees, are included in this dataset. It is a challenging dataset because it only contains limited number of samples and their variations are complex. We have also tested our method on a recent light field dataset~\cite{pratul2017lightField} of flowers (Flower dataset). This dataset contains 3433 light field images of various kind of flowers. We randomly split flower dataset as \cite{pratul2017lightField}, getting 3233 for training and 100 for testing. For the trade-off in time and space requirement of the network, we randomly crop the original input image from $541\times376$ to $320\times240$ for training. We use a mini-batches of $4$ to have the best trade-off between speed and convergence. In experiments, our network is trained in 12k iterations. The whole experiment takes almost $2$ days for training. Following the work of \cite{LearningViewSynthesis}, the weights of our network are initialized by Xavier \cite{glorot2010understanding} approach. We use ADAM \cite{kingma2014adam} for optimization, with $\beta_1$ = 0.9, $\beta_2$ = 0.999 and learning rate of 0.0001.





\section{Experiments}

In this section, we first compare our method with some baseline works on VS100 dataset.

Besides~\cite{pratul2017lightField}, we adapt the state-of-the-art methods for stereo pair synthesis\cite{xie2016deep3d}, for synthesis with multi-view input\cite{LearningViewSynthesis} and for handling single object\cite{zhou2016view}, to fit our problem to perform the comparisons, because there are few previous works focusing on synthesizing user-specified novel viewpoints from one single nature image.
As \cite{pratul2017lightField} is our most relevant work, we also compare with it on the Flower dataset, which is used in their experiments.
Several evaluation metrics, such as Peak Signal to Noise Ratio (PSNR), structural similarity (SSIM) and Mean Absolute Error (MAE), are used for the quantitative comparisons. And image and video results are also demonstrated in our paper and supplementary materials.

Then, to evaluate the effectiveness of our key contributions, i.e. the depth-assistant mechanism and the full resolution network, we further test our method by removing them or replacing them with traditional solutions.
The experiments show that either of them is crucial to the final good results of our technique.
Next, we show more results of our technique, where unconstrained images from the Internet are also tested.
Finally, we demonstrate the application of our method on image refocusing and discuss the limitations of our technique.
\begin{figure*}[htb]
\setlength{\abovecaptionskip}{-0.0cm}
\setlength{\belowcaptionskip}{-0.4cm}
\centering
\includegraphics[width=\textwidth]{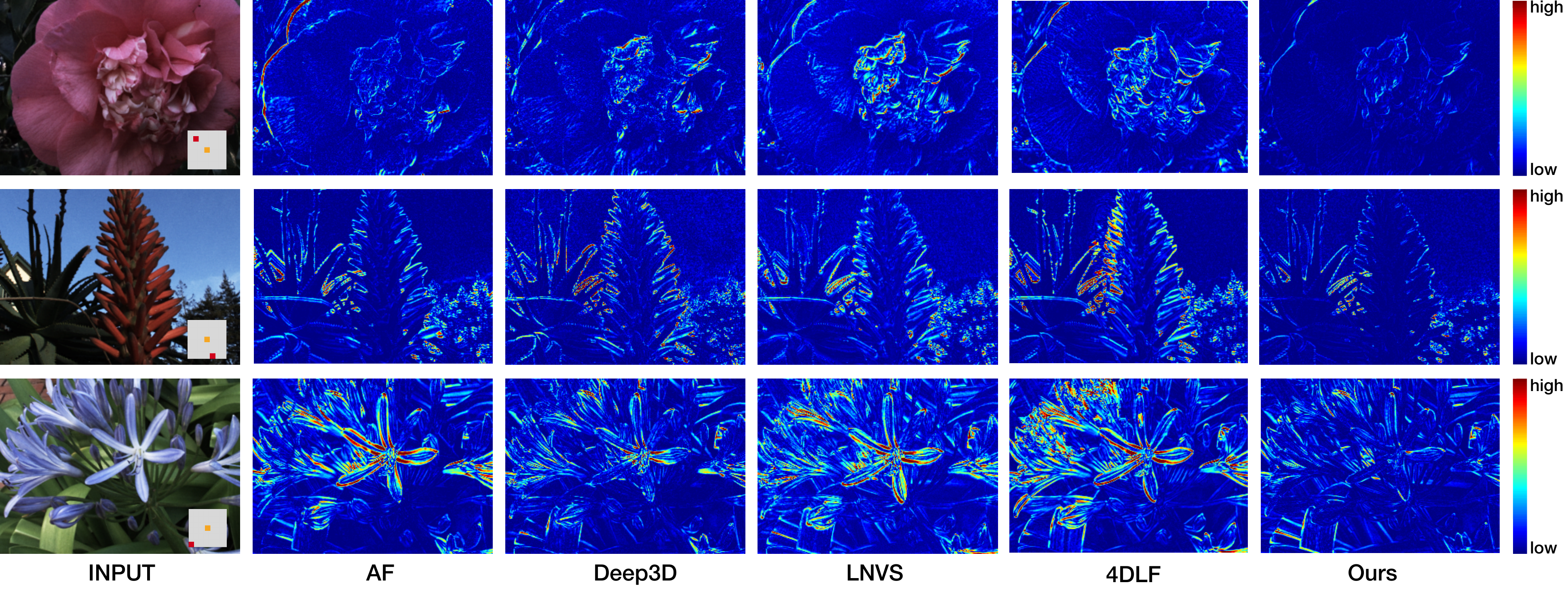}
\caption{Image reconstruction errors for different scenes in different viewpoints. Compared with the other methods, our approach shows the minimum reconstruction errors. }\label{fig:4}
\end{figure*}

\subsection{Baseline methods}
\vspace{-3mm}

Here we introduce the necessary adaptations on the baseline methods.

\textbf{Deep3D} \cite{xie2016deep3d} predicts the right view image from a single left view image. 
Considering the GPU memory, we use $100$ disparity layers in total, and re-sample the disparity range with $10$ disparity levels on either the horizontal and the vertical direction. For the network, we follow the original Deep3D method. Features, which are extracted by pre-trained VGG16 \cite{Simonyan14c} network, are used for processing the input view. Disparity candidates are rebuilt by connecting the re-sampled high-level VGG16 features and other low-level features.

\textbf{LNVS} \cite{LearningViewSynthesis} proposes a network to use four corner views to predict novel views in-between. 
By replacing the four corner input views to one center image, we extend this method to our problem. Because there is only one view as the input, other than the mean and variance of input views, we just feed the $100$ disparity levels as the $100$ layer features to the neural network.


\textbf{AF} \cite{zhou2016view} focuses on novel view synthesis for an object and fly-through a scene. We extend this method by replacing the transformation parameters of objects to viewpoint coordinates directly.

\textbf{4DLF} \cite{pratul2017lightField} proposes a network for light field reconstruction for flowers. 
To compare this method with our method on a general dataset, we pre-train their network on the Flower dataset as they proposed, and then fine-tune the network to fit VS100 dataset at a lower learning rate. The network trained directly on VS100 is also compared in Table \ref{tab:2}, denoted as \textit{4DLF(VS100)}.

\begin{table}[htbp!]
\caption{Numerical comparison of our method and the state-of-art methods. A larger value indicates better quality for PSNR and SSIM, so does a less value for MAE.}
\label{tab:2}
\setlength{\abovecaptionskip}{-0.0cm}
\centering
\begin{tabularx}{\columnwidth}{XXXX}
\hline
                & PSNR $\uparrow$    & SSIM $\uparrow$  & MAE $\downarrow$   \\ \hline
4DLF(VS100)              & 33.0853     & 0.8299    & 0.0328     \\
4DLF             & 34.5788 & 0.8545 & 0.0285  \\
LNVS            & 34.1789 & 0.8483 & 0.0282 \\
Deep3D          & 34.9809 & 0.8567 & 0.0232 \\
AF              & 35.5367 & 0.8531 & 0.0237 \\
Ours            & \textbf{36.4401} & \textbf{0.8875} & \textbf{0.0202}  \\ \hline
\end{tabularx}

\end{table}

\subsection{Comparisons on VS100 dataset }

We test our method and the aforementioned four methods on all the 30 test images in the VS100 dataset and generate 48 novel viewpoints for each image. By comparing with the ground truth, we calculate three numeric metrics and represent the average values in Table \ref{tab:2}. We can clearly see that our approach outperforms all the other methods. Notice that except for our method, all the other ones do not need the depth order dataset to train a depth predictor, but we treat this as a contribution of our technique as it really benefits to the task of single view-based viewpoint synthesis and it only requires a pre-training step which can be performed in advance and does not need to be changed in all the following steps.


Besides the quantitative comparisons, we also show some synthesized results in Figure \ref{fig:3}. 

For Deep3D, as the sampled disparities are quite sparse. It is difficult to assign correct disparities for all patches in images. As shown in the result in the second row, a small part of the building is missed as its disparity is not correctly estimated. For LNVS, without the constraints from the four corner images, the cascade network generates blurring results. The same problem happens in 4DLF. As it tries to learn by patches ($192 \times 192$), it is more suitable for constrained scenes with large number of training samples, such as the Flower dataset. VS100 dataset is more challenging as it contains various different natural scenes but a limited number of samples, so it does not perform well in this situation. For our method, as we have better depth information to infer disparity, we get noticeably better results. AF predicts the flow of appearance by an encoder-decoder structure. As illustrated before, this structure will not preserve pixel level details for image transformations, and thus generates artifacts (see the edge of the building is no longer straight in the result of the second row).

The reconstruction errors between the ground truth and the synthesized novel view are also shown in Figure \ref{fig:4}, and our method shows the minimum reconstruction errors than the other methods both in the views near (top row) the input and in the corner views (bottom row). A more clear demonstration for comparing the five methods is shown in the video.

\begin{figure} [tbp!]
\setlength{\abovecaptionskip}{0.cm}
\setlength{\belowcaptionskip}{-0.3cm}
\centering
\includegraphics[width=0.95\columnwidth]{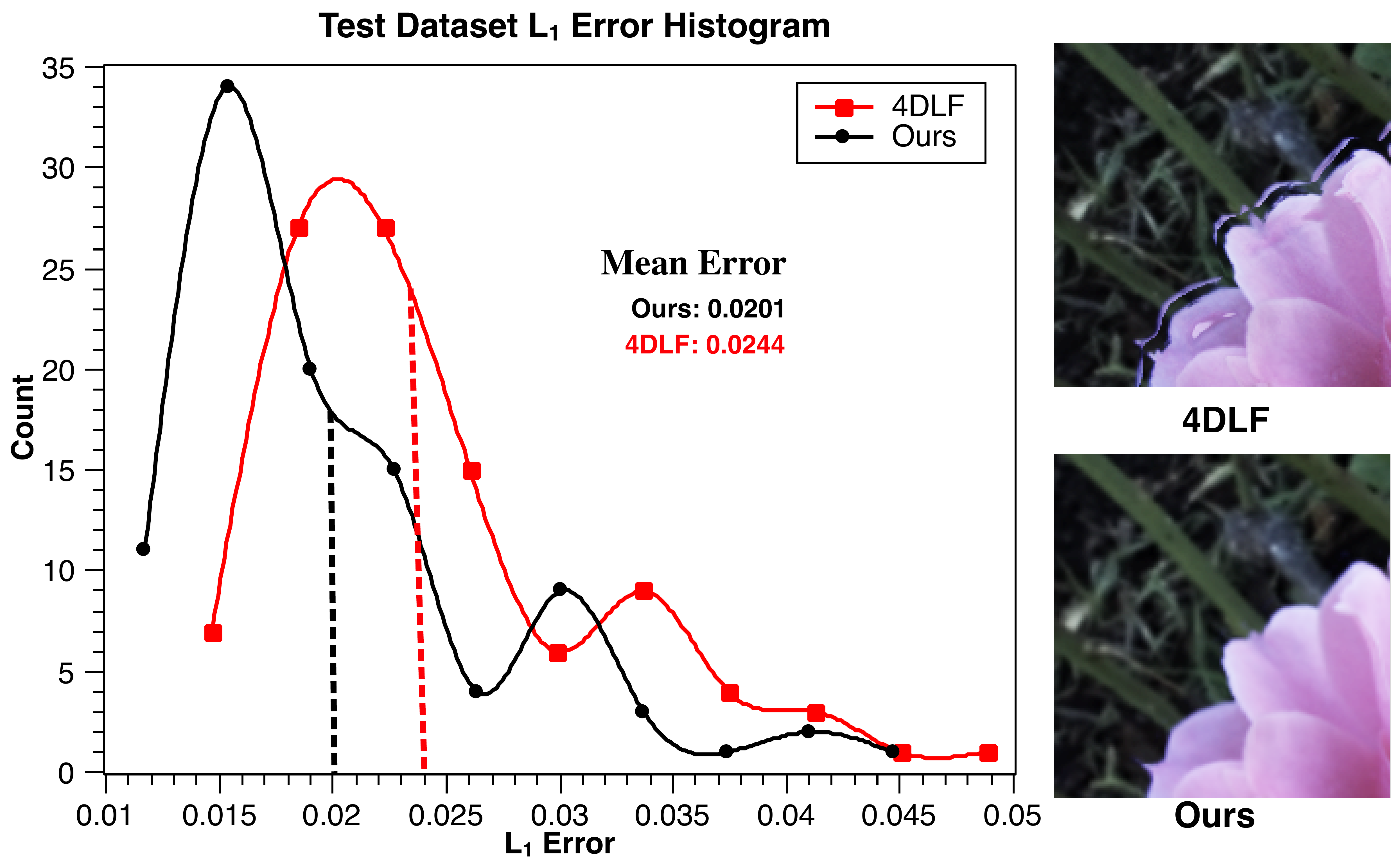}
\caption{Comparison between 4DLF method and our method on the test set of Flower dataset. Left histogram plot is the MAE comparison between 4DLF and our method. The mean MAE of our method is much lower than 4DLF. As for the visual results on the right, as the second occlusion network of 4DLF fails in some examples, the border of the flower is not as good as our result. This difference is better viewed by zooming in the figure.
}
\label{fig:his}
\end{figure}
\subsection{Comparisons on Flower dataset }

We also compare our method and 4DLF\cite{pratul2017lightField} on the Flower dataset. Our method is trained on the $320\times240$ random crops of the original image, while we train \cite{pratul2017lightField} as reported in their paper. To fit our image ratio, we center crop the original test images from $542\times364$ to $480\times360$ to perform the comparison. For our method, we resize the test image to $320\times240$ to fit our model. And we resize the output results back to $480\times360$ for comparison. For \cite{pratul2017lightField}, we follow their methods to test the images on the same crop of $480\times360$. Figure \ref{fig:his} shows the histogram of MAE error on the test dataset. The MAE of our method mainly falls in the interval of $[0.012,0.023]$ where their MAE error most falls in $[0.015,0.028]$. The mean MAE error of our method ($0.0201$) is also better than their method ($0.0244$). Figure \ref{fig:his} also shows the failure case of \cite{pratul2017lightField} in large occlusion regions. As large occlusions are caused by large depth discontinuity and their results are sensitive to depth errors when synthesizing novel viewpoints, the failure case indicates that 3D-CNN and dilated convolution used in \cite{pratul2017lightField} cannot generate very accurate depth in these regions. More comparisons of our method and 4DLF can be found in the supplementary figure.

\begin{figure} [tbp!]
\setlength{\abovecaptionskip}{0.cm}
\setlength{\belowcaptionskip}{-0.3cm}
\centering
\includegraphics[width=0.95\columnwidth]{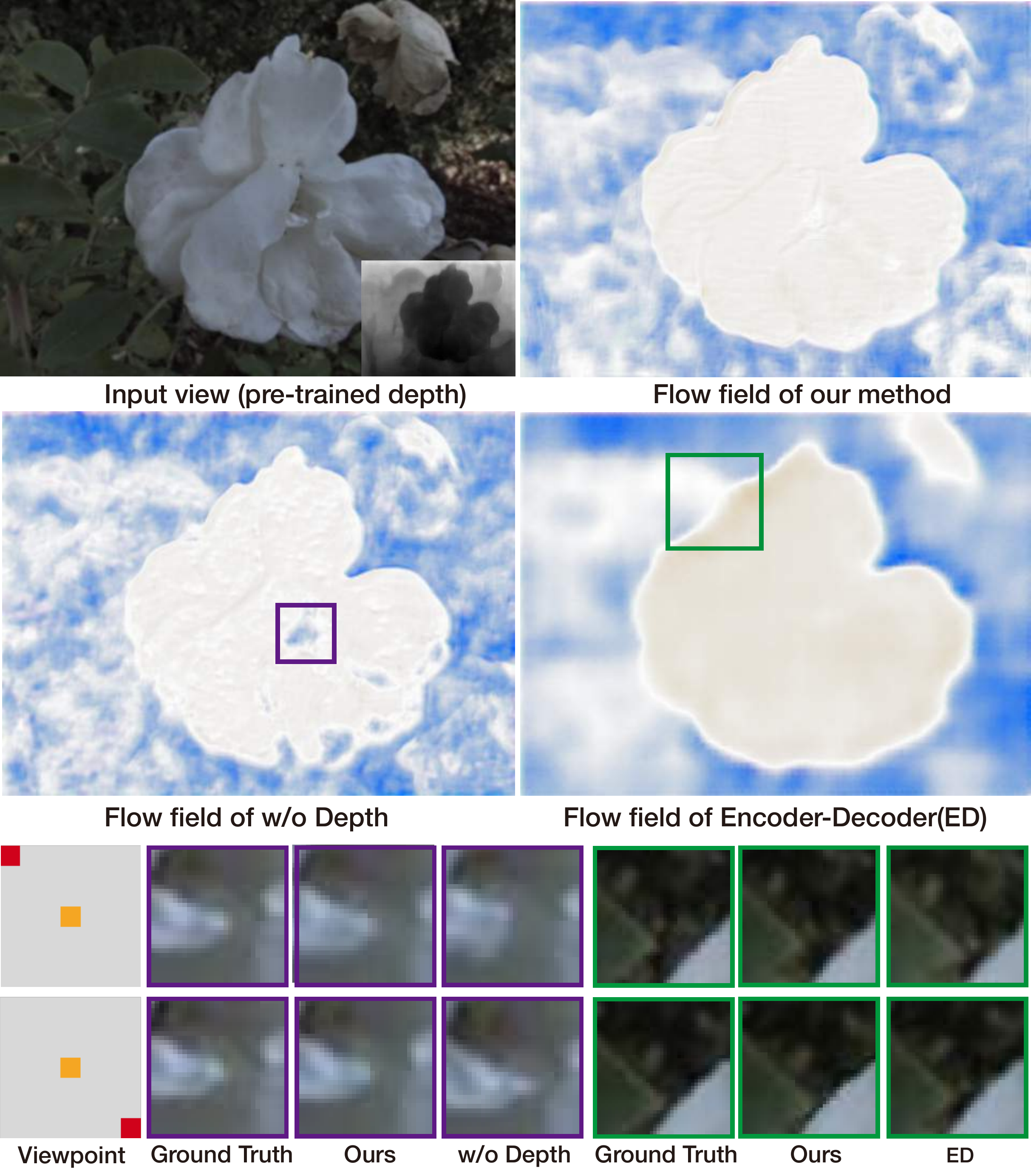}
\caption{Evaluation of the key techniques of our method.
The top part shows the color-coded flow field estimated by different alternative solutions, while the lower part shows the corresponding synthesized results.
For the method w/o depth, the center of a flower is regarded as background, leading to the incorrect transformation of the flower center, shown in the purple rectangle.
For the Encoder-decoder method, the flow looks blurring as local features (such as edges) are missing, leading the background to be translated as foreground, shown in the green rectangle.
}

\label{fig:5}
\end{figure}

\vspace{-1.5mm}
\subsection{Evaluations}
We first evaluate the full resolution network of our method.
We believe this network is more suitable for our task of dense pixel value synthesis, compared with the traditional Encoder-Decoder network which is widely used for tasks with sparse outputs, like \cite{long2015fully} and \cite{zhou2016view}.
As shown in Table \ref{tab:3} and Figure \ref{fig:5}, our full resolution network performs better in our task.
The underline reason is that the traditional encoder and decoder network basically represents the input image as low-resolution features with high dimension.
As features are already in low resolution, it is difficult to reconstruct output images with highly accurate pixel positions.
From the green crop of Figure \ref{fig:5}, because of the blurred flow field, the background leaf transforms with the foreground flower while our method shows the correct movement similar to the ground truth.



Then we evaluate the effectiveness of involving depth information in our system. We know that depth is important because it actually determines the disparity of scene objects in different viewpoints. However, as depth estimation from a single input is an ill-posed problem, we utilize a network trained on a very large image dataset to get a reference depth as good as possible. And as image warping can be fully decided by local depth information, our synthesis network do not need to focus on global image features when the depth is already encoded in a feature layer. So involving depth matches our full resolution network which mainly focuses on local features. From Table \ref{tab:3} and Figure \ref{fig:5}, the network without predicted depth considers the border of flower and the center of flower as background by their local colors, generating artifacts indicated in the purple patch.

\begin{table}[hb]
\centering
\setlength{\abovecaptionskip}{-0.0cm}
\setlength{\belowcaptionskip}{-0.2cm}
\caption{Numerical comparison of our method and some alternative solutions. }
\label{tab:3}
\begin{tabularx}{\columnwidth}{lccc}
\hline
                            & PSNR $\uparrow$    & SSIM $\uparrow$  & MAE $\downarrow$      \\ \hline
Using Encoder-Decoder       				& 35.6235          & 0.8648          & 0.0215         \\
W/O Predicted Depth         		& 35.5281             & 0.8624            & 0.0224         \\
Ours Full 					& \textbf{36.4401} & \textbf{0.8875} & \textbf{0.0202}  \\ \hline
\end{tabularx}

\end{table}

\begin{figure}[tbp!]
\centering
\includegraphics[width=\columnwidth]{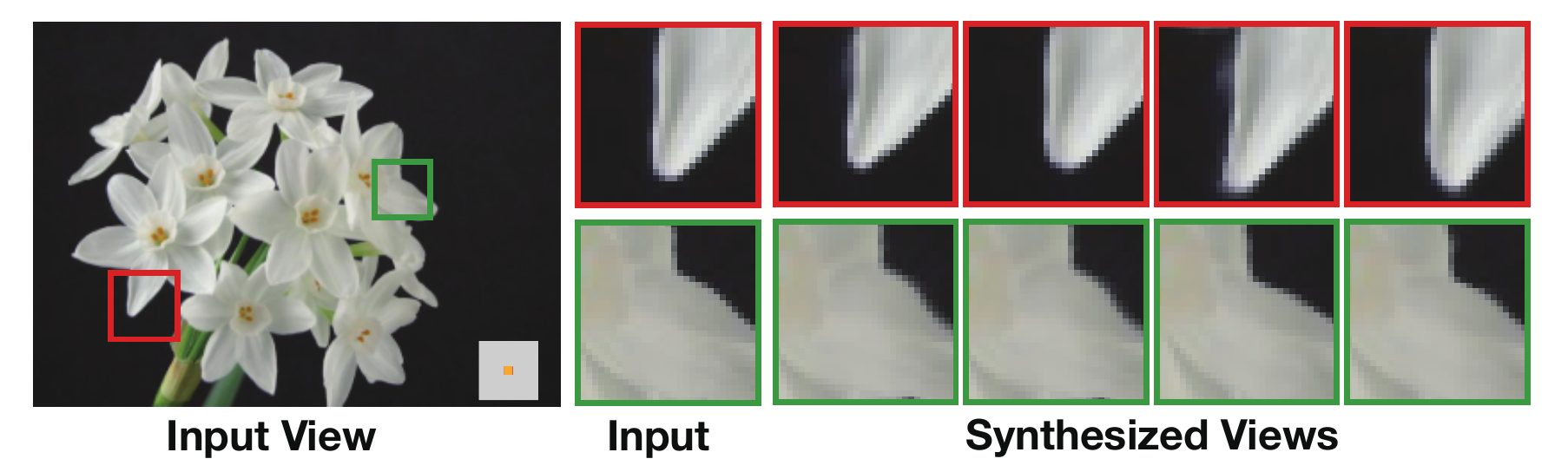}
\caption{ Although VS100 dataset is limited, our method still shows good generalization capability to normal scenes. This is an example of our method performed on a random image from the Internet. We treat this image at the $(0,0)$ coordinates of the light field, and feed it to our network to synthesis the surrounding viewpoints. The right images show the comparisons between the selected patches of the input view and the four synthesized corner views at $(-3,-3)$, $(-3,3)$, $(3,-3)$ and $(3,3)$, respectively. }
\label{fig:more}
\end{figure}


\vspace{-1.5mm}
\subsection{More results and Image Refocusing}

Our method can handle images of various scenes, not only the test images in the dataset, but also the Internet images.
All our results are shown in the supplementary video for a better demonstration. some example are shown in Figure \ref{fig:0} and \ref{fig:more}, where local regions are correctly changed with the change of the viewpoint.

\begin{figure} [tbp!]
\setlength{\abovecaptionskip}{0.cm}
\setlength{\belowcaptionskip}{-0.3cm}
\centering
\includegraphics[width=0.95\columnwidth]{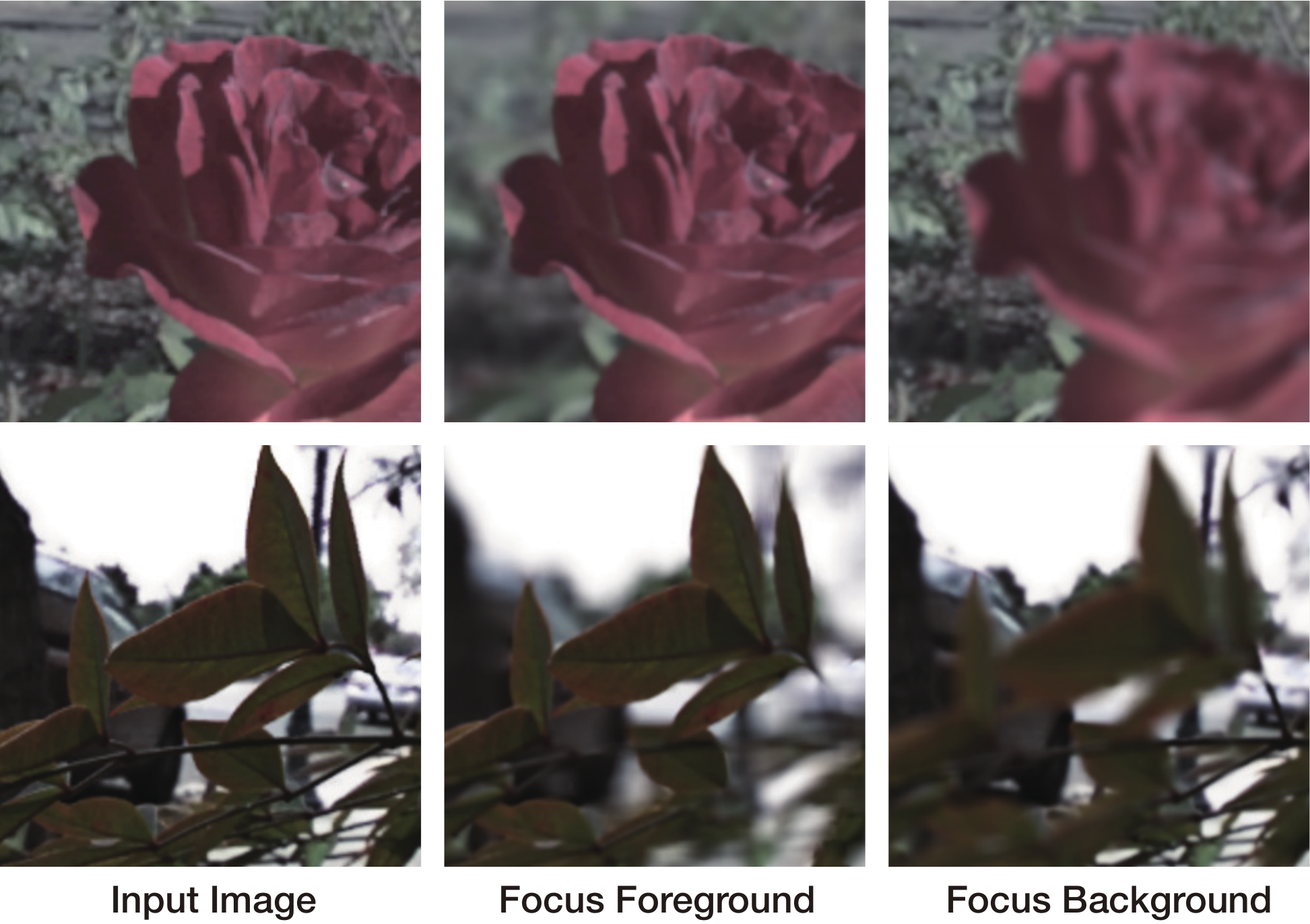}
\caption{Some image refocusing examples of our method. The top example is from the test set of Flower dataset while the bottom example is from the test set of VS100 dataset. From left to right are the original input image, the image focusing on foreground and the image focusing on background.
}
\label{fig:focus}
\end{figure}

As our method can also be regarded as a light field reconstruction technique from a single image (by synthesizing dense viewpoints surrounding the input), it is possible to refocus a 2D image by adding all the light field images together with different offsets, as shown in Figure \ref{fig:focus}.

\vspace{-1.5mm}
\subsection{Limitations}
First, we can not handle nature images of arbitrary scenes, as there are many scenes never being seen by our light field dataset. But we argue that we do not need a large amount of samples of a particular scene, as we only need to train a network for local feature extraction. Second, our full resolution network requires more memory cost and training/testing time compared with the traditional encoder-decoder network. Third, we can not synthesize novel viewpoint which is far from the input one because it requires much more information which is missing in the input.

\section{Conclusion}

In this paper, we propose a method to synthesize user-desired novel views from one single input image. It is challenging and ill-posed, and difficult for current powerful deep learning techniques as there does not exist sufficient light field dataset for training. To tackle this problem, we first leverage a large image dataset with sparsely labeled depth orders to train a depth predictor. We demonstrate that combining the depth with only the local image features extracted by a specially designed full resolution network, novel view synthesis can be achieved on various input images. As the full resolution network only needs to extract local features, the current light field dataset is sufficient as shown in our experiments on the VS100 dataset. So we have made a step further in increasing the generalization capability on this new and important task.

{\small
\bibliographystyle{ieee}
\bibliography{egpaper_for_review}
}

\end{document}